\documentclass[journal]{IEEEtran}

\usepackage{ifpdf}
\usepackage{xcolor}

\ifCLASSINFOpdf
  
\else
  
\fi

\usepackage{graphicx}
\graphicspath{{./img/}}

\usepackage{times, epsfig,rotating}
\usepackage{amsmath,amsthm,amssymb}
\usepackage{mathtools}
\usepackage{subfigure}

\usepackage[hidelinks]{hyperref}

\usepackage{lipsum}
\usepackage{graphicx}
\usepackage{multirow}
\usepackage{algorithm}
\usepackage[noend]{algpseudocode}
\usepackage{array}
\usepackage{fixltx2e}
\usepackage{stfloats}
\usepackage{slashbox}
\usepackage{pdflscape}
\usepackage{afterpage}
\usepackage{xcolor}
\usepackage{caption}

\usepackage{adjustbox}
\usepackage{fancyhdr}

\fancypagestyle{firstpage}{%
  \lhead{This work has been submitted to the IEEE for possible publication. Copyright may be transferred without notice, after which this version may no longer be accessible.}
}

\newtheorem{mydef}{Definition}

\newtheorem{theorem}{Theorem}

\begin{document}


\title{PIDA: Smooth and Stable Flight Using Stochastic Dual Simplex Algorithm and Genetic Filter}


\author{Seid~Miad~Zandavi,
		Vera~Chung,
		Ali~Anaissi
\thanks{Seid M. Zandavi, Vera Chung and Ali Anaissi are with the School of Computer Science, The University of Sydney, Sydney NSW 2006 Australia e-mails: \{miad.zandavi, vera.chung, ali.anaissi\}@sydney.edu.au.}
}



\maketitle
\thispagestyle{firstpage}

\begin{abstract}
This paper presents a new Proportional-Integral-Derivative-Accelerated (PIDA) control with a derivative filter to improve quadcopter flight stability in a noisy environment. The mathematical model is derived from having an accurate model with a high level of fidelity by addressing the problems of non-linearity, uncertainties, and coupling. These uncertainties and measurement noises cause instability in flight and automatic hovering. The proposed controller associated with a heuristic Genetic Filter (GF) addresses these challenges. The tuning of the proposed PIDA controller associated with the objective of controlling is performed by Stochastic Dual Simplex Algorithm (SDSA). GF is applied to the PIDA control to estimate the observed states and parameters of quadcopters in both attitude and altitude. The simulation results show that the proposed control associated with GF has a strong ability to track the desired point in the presence of disturbances. 

\end{abstract}

\begin{IEEEkeywords}
Drone, Control, PIDA, SDSA, Genetic Filter.
\end{IEEEkeywords}



\section{Introduction}

\IEEEPARstart{C}{ontrol} is exciting areas of research in self-driving/autonomous system, as many engineering applications required this factor. Recently, unmanned aerial vehicles (UAVs) have gained the attention of many researchers working in different applications, such as search and rescue, delivery, and crowdsourcing \cite{kim2019adaptive, koh2012dawn}. UAVs or drones have been developed in many areas, including robotics, control, path planning, and communication \cite{phung2017enhanced, rajappa2016adaptive, derafa2012super,woods2017novel}. The current attention to increasing the usability of drones in many commercial and civil applications inspires researchers to make this dynamic system more controllable. In particular, quadcopters are popular drones due to their performance in terms of vertical take-off, landing, their simple and stable structures. However, their instability, unstable dynamics, non-linearity, and cross-coupling make this system an interesting underactuated system \cite{meng2018distributed}. Generally, a quadcopter has six degrees of freedom, although four rotors should control all directions. This causes cross-coupling between rotation and translational motions \cite{santoso2018robust}. Therefore, the nonlinear dynamics needs to be managed by the controller. 

In the past, classical control techniques were applied to address autonomous formulations. The main issue is to formulate an accurate model describing a dynamic system. This means that any changes and modifications in the system, such as uncertainties in both model and environment, affect the performance of the controller, making it necessary to update the controller parameters. Thus, when the system’s dynamics and its operation through the environment change, its control parameters and features require re-tuning \cite{milovanovic2020adaptive, gu2019pid,puchta2019optimization}.

Various control algorithms have been developed to manage the non-linearity of the quadrotor. For example, command-filtered Proportional-Derivative (PD) \cite{santoso2019hybrid} or Proportional-Integral-Derivative (PID) control \cite{zuo2010trajectory}, integral predictive control \cite{raffo2010integral}, and optimal control \cite{ritz2011quadrocopter, zandavi2018multidisciplinary} have been applied. The Sliding Mode Control (SMC) is another common control algorithm that is used to improve performance in terms of stability due to the influence of modeling errors and external disturbances \cite{derafa2012super,xu2006sliding,besnard2007control,wen2016fuzzy}. Note that the chattering effect in the SMC arises in the steady state, where it simulates unmodeled frequencies of the system dynamics.

Of these controllers, PID is preferred due to its simplicity and ability to adapt to unknown changes. These fast and simple features make the PID control strategy efficient and versatile in robotics, although it causes wide overshoot and large settling time \cite{ang2005pid}. Initially, the parameters of a PID controller, called gains, were set by the expertise in obeying certain rules, such as investigating its step responses, Bode plots and Nyquist diagrams. However, when the complex environment changes and affects the dynamics of the quadcopter, the PID parameters must to be re-tuned and it is essential to consider uncertainties in formulating the model. In this regard, uncertainties and stochastic processes in the dynamic system can be modeled as color noise and white noise. Thus, the derivative term is able to cope the effect of disturbances.

To manage unstable systems, derivative plays a significant role in improved control loop performance. Mathematically, the derivative terms in PID controller open an avenue for more actions when the error (i.e., following the desired response) is fluctuating wildly. Thus, an additional derivative term (i.e., zero) can decrease the size of overshooting \cite{jung1996analytic}. This can improve controllability. Additionally, this derivative term supports a better response in terms of speed and smoothness, where limiting overshoot and settling time in an acceptable bound are considered.

In addition to the control, integrated estimation of states and parameters plays an important role in improving the performance of the quadcopter in the presence of uncertainties and measurement noise \cite{yuan2018nonfragile}. Two different categories of classical and heuristic \cite{zandavi2019state} filters have been used to address the state estimation problem. For example, Kalman Filter (KF) \cite{kalman1960new}, Extended Kalman Filter (EKF) \cite{jazwinski2007stochastic}, Unscented Kalman Filter (UKF) \cite{julier1997new} are classical filters, and Particle Filter (PF) \cite{carpenter1999improved}, Simplex Filter (SF) \cite{nobahari2016simplex} and Genetic Filter (GF) \cite{zandavi2019state} are introduced as heuristic filters \cite{zandavi2019state}.

Heuristic filters work based on point mass (or particle) representation of the probability densities \cite{arulampalam2002tutorial}. Unlike UKF, PF represents the required posterior Probability Density Function (PDF) by a set of random samples instead of deterministic ones. Also, it uses a resampling process to reduce the degeneracy of particles. The standard resampling process copies the important particles and discards insignificant ones based on their fitness. This strategy suffers from the gradual loss of diversity among the particles, known as sample impoverishment. Researchers have proposed different resampling strategies such as Binary Search \cite{gordon1993novel}, systematic resampling \cite{smith2013sequential} and residual resampling \cite{arulampalam2002tutorial}. Some heuristic optimization algorithms have also been inserted to PF to improve its performance. For example, SF \cite{nobahari2016simplex} utilizes Nelder-Mead simplex approach for state estimation. GF \cite{zandavi2019state} was utilized a genetic algorithm scheme, and its operators to estimate the state of dynamic systems.

In this paper, the new accelerated PID controller with derivative filter associated with GF is proposed to make an unstable quadcopter track the desired reference with the proper stability. GF is utilized to estimate the height and vertical velocity of the modeled dynamic system (i.e., the quadcopter) while hovering. Consequently, the mathematical model of the dynamic system is provided and considers non-linearity, instability, cross-coupling among different modes (i.e., pitch, roll, and yaw), and the uncertain environment. The controller parameters are tuned using the Stochastic Dual Simplex Algorithm (SDSA) optimization algorithm \cite{ZandaviSDSA2019}, which improves the trade-off between exploration and exploitation to achieve better optimal parameters for the proposed controller. 

This paper is organized as follows. Section \ref{Control_sec2} describes the mathematical model of the dynamic system. The proposed controller is introduced in Section \ref{Control_sec3}. Stability analysis is presented in Section \ref{Control_sec4}. Optimization and heuristic filter are explored in Section \ref{OptiHeuFil}. Numerical results and discussion are given in Section \ref{Control_sec5}. Finally, the paper ends with the conclusion in Section \ref{Control_sec6}. 


\section{Dynamic Model}\label{Control_sec2}
\begin{figure}[h]
    \centering
    \includegraphics[trim ={8.5cm 15.5cm 4.0cm 4.5cm},clip,scale = 1]{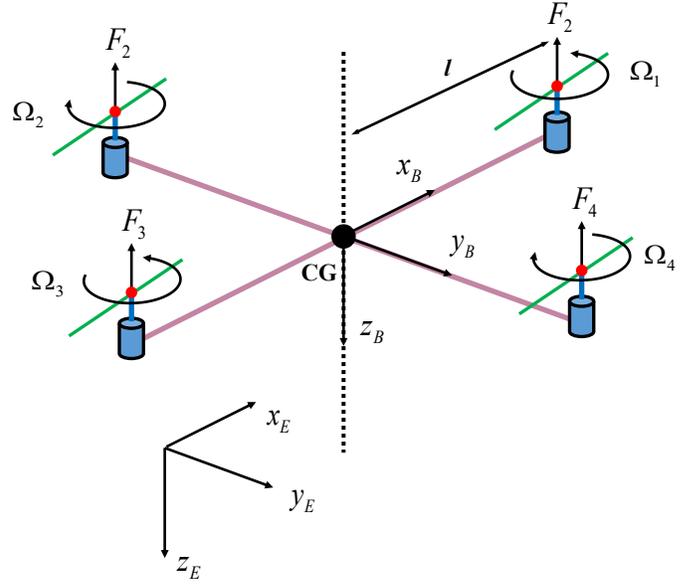}
    \caption{Earth Fixed and Body Fixed coordinate systems}
    \label{fig_Control1}
\end{figure}

The mathematical model of a system can be used as the first step to study its performance. In this regard, the quadcopter studied in this paper is modeled in Fig. \ref{fig_Control1}, considering earth-centered inertia (ECI) and body frame. Thus, $X_E = [x_E , y_E , z_E]^T$ and $X_B = [x_B , y_B , z_B]^T$ are defined as transformational motions from inertia frame to body frame due to having an accurate dynamic model.


The attitude of the quadcopter is formulated based on the Eular angles roll, pitch, and yaw, which are rotated from the x-axis, y-axis and z-axis, respectively. Thus, the Eular angles are $\Theta = [\phi, \theta, \psi]^T$, and the angular velocity in the body frame is $\dot{\Theta} = [\dot{\phi}, \dot{\theta}, \dot{\psi}]^T$. In this sense, the angular velocity in inertia ($\omega = [p, q, r]^T$) is formulated as follows:

\begin{equation}
\label{Control_eq1}
	\omega =  \left[ {\begin{array}{ccc}
									1 & 0 & -\sin(\theta) \\
									0 & \cos(\phi) & \cos(\theta) \sin(\phi) \\
									0 & -\sin(\phi) & \cos(\theta) \cos(\phi)  \end{array}} \right] \cdot \dot{\Theta}
\end{equation}

Total torques are caused by three segments: thrust forces ($\tau$), body gyroscopic torque (${\tau}_b$) and aerodynamic friction (${\tau}_a$). In addition, each component of the torque vector ($\tau = [{\tau}_{\phi},{\tau}_{\theta},{\tau}_{\psi}]^T$), corresponding to a rotation in the roll, pitch, and yaw axis, can be determined by Eqs (\ref{Control_eq2})–(\ref{Control_eq4}):

\begin{equation}
\label{Control_eq2}
{\tau}_{\phi} = l (F_2 - F_4)
\end{equation}

\begin{equation}
\label{Control_eq3}
{\tau}_{\theta} = l (F_3 - F_1)
\end{equation}

\begin{equation}
\label{Control_eq4}
{\tau}_{\psi} = c (F_2 - F_1 + F_4 - F_3)
\end{equation}
where \textit{l} is the distance between the center of motor and the center of mass, and \textit{c} is the force to torque coefficient. As assumed, the quadcopter is a rigid body and symmetrical dynamics apply, from which the torque can be calculated by following equation: 

\begin{equation}
\label{Control_eq5}
{\tau} = I \dot{\omega} + \Omega (I \omega)
\end{equation}

where \textit{l} is the distance between the center of motor to center of mass, and \textit{c} is the force to torque coefficient. As assumed, the quadcopter is a rigid body and symmetrical dynamics, from which the following equation can calculate the torque: 

\begin{equation}
\label{Control_eq6}
{\Omega} = \left[ {\begin{array}{ccc}
									0 & -r & q \\
									r &  0 & -p \\
									-q & p & 0 \end{array}} \right]
\end{equation}

In this system, the main control inputs are correlated to the torque ($\tau = [{\tau}_{\phi},{\tau}_{\theta},{\tau}_{\psi}]^T$) caused by thrust forces, body gyroscopic effects, propeller gyroscopic effects and aerodynamic friction. Gyroscopic effects and aerodynamic friction are considered external disturbances for the control. Thus, control inputs are determined as Eq (\ref{Control_eq7}).

\begin{equation}
\label{Control_eq7}
\left[ {\begin{array}{c}
	u_{\phi} \\ u_{\theta} \\ u_{\psi} \\ u_{T} 
	\end{array}} \right]
	=
	\left[ {\begin{array}{c}
	\tau_{\phi} \\ \tau_{\theta} \\ \tau_{\psi} \\ \tau_{T} 
	\end{array}} \right]
	=
	\left[ {\begin{array}{cccc}
		 0 & l & 0 & -l \\
		-l & 0 & l & 0 \\
		-c & c & -c & c \\
		 1 & 1 & 1 & 1
		\end{array}} \right]
		\left[ {\begin{array}{c}
			 F_1 \\ F_2 \\ F_3 \\ F_4 
		\end{array}} \right]
\end{equation}
where $\tau_{T}$ is the lift force and $u_T$ corresponds to the total thrust acting on the four propellers, where $u_{\phi}$, $u_{\theta}$ and $u_{\psi}$ represent the roll, pitch, and yaw, respectively. The drone’s altitude can be controlled by lift force ($u_T$), which is equal to quadcopter weight. The dynamic equations of the quadcopter are formulated based on the Newton-Euler method \cite{zipfel2007modeling}. The six degree of freedom (6-DOF) motion equations are stated by Eqs (\ref{Control_eq8})–(\ref{Control_eq13}).

\begin{equation}
\label{Control_eq8}
\dot{u} = rv -qw -g \sin(\theta)
\end{equation}

\begin{equation}
\label{Control_eq9}
\dot{v} = pw -ru +g  \sin(\phi) \cos(\theta)
\end{equation}

\begin{equation}
\label{Control_eq10}
\dot{w} = qu -pv + g \cos(\theta) \cos(\phi) - \frac{1}{m} u_T
\end{equation}

\begin{equation}
\label{Control_eq11}
\dot{p} = \frac{1}{I_{xx}} \left[ (I_{yy}-I_{zz})qr + u_{\phi} + d_{\phi} \right]
\end{equation}

\begin{equation}
\label{Control_eq12}
\dot{q} = \frac{1}{I_{yy}} \left[ (I_{zz}-I_{xx})pr + u_{\theta} + d_{\theta} \right]
\end{equation}

\begin{equation}
\label{Control_eq13}
\dot{r} = \frac{1}{I_{zz}} \left[ (I_{xx}-I_{yy})pq + u_{\psi} + d_{\psi} \right]
\end{equation}
where $d = [d_{\phi},d_{\theta},d_{\psi}]^T$ is the angular acceleration disturbance corresponded to propeller angular speed, and these acceleration disturbances are modeled by Eq (\ref{Control_eq14}).

\begin{equation}
\label{Control_eq14}
 d = \left[ {\begin{array}{c}
	+  qI_{m}  \Omega_{r} \\ -pI_{m} \Omega_{r} \\ 0 
	\end{array}} \right]
\end{equation}
where $\Omega_r = \sum_{i=1}^{4} (-1)^{i} \Omega_i $ is the overall residual propeller angular speed, and $\Omega_i$ is the angular velocity of each rotor. $I_{m}$ is the rotor moment of inertia around the axis of rotation. Hence, the dynamics equations of the system can be summarized as follows: 

\begin{align}
\label{Control_eq15}
\begin{split}
\dot{x}(t) = A(x) + B(x)u(t) + d  \\
y(t) = C(x) + D(x)u(t)
\end{split}
\end{align}

where $x = [\phi, \theta, \psi, p, q, r, w]^T$ and $y = [y_1,y_2,y_3,y_4]^T$ are the states and measurable outputs, respectively. $u = [u_1,u_2,u_3,u_4]^T$ is the control and $d$ is the disturbance. $A$, $B$, $C$, and $D$ are the nonlinear functions regarding dynamic equations of the system. 

The control design is considered to minimize the error for tracking the desired command (see Eq (\ref{Control_eq16})).

\begin{equation}
\label{Control_eq16}
\lim_{t \to \infty}{\|{e(t)}\|} = \varepsilon
\end{equation}
where $e(t) = r(t)-y(t)$ is the difference between reference inputs and the system’s measurable outputs and $\varepsilon$ is the small positive value.


\section{Proposed PIDA Controller} \label{Control_sec3}

The PID control is applied to many engineering applications because of its simplicity. Note that PID cannot function effectively when wide overshoot and considerable settling time occur in the system. A modified PID controller can address this issue by adding a zero known as PID acceleration (PIDA). It is employed to achieve a faster and smoother response for a higher-order system and retains both overshoots and settling time within an acceptable limit. The proposed linear control can also control the nonlinear system. In this approach, the dynamic airframe is linearized about the equilibrium point. The linearization of the model is given by Eq (\ref{Control_eq17}).

\begin{equation}
\label{Control_eq17}
\Delta \dot{X} = J_X \Delta X + J_U \Delta U 
\end{equation}
where $J_X$ and $J_U$ are the Jacobian transformation of the nonlinear model about the equilibrium point ($X_{eq} = [\phi_0,\theta_0,\psi_0,p_0,q_0,r_0,w_0]^T$). Note that the equilibrium point can be calculated by solving $\dot{X} = AX = 0$. Any solution can be the equilibrium point because of the null space if $det(A)$ is equal to zero.

In this regard, a multi-input, multi-output (MIMO) control system design follows the desire command in altitude and attitude channels. A MIMO tracking controller can not only stabilize the system, but also make it follow a reference input. Thus, the linear system is given as follows: 

\begin{equation}
\label{Control_eq18}
{\begin{array}{c}
\dot{X} = AX + BU + D_d \\
   Y = CX
	\end{array}}
\end{equation}
where $Y$ is the outputs that follow the reference inputs and $D_d = [0, 0, 0, d^T, 0]^T$ is the angular disturbance. In this approach, the integral state is defined as follows: 

\begin{equation}
\label{Control_eq19}
\dot{X}_N = R-Y = R-CX 
\end{equation}

According to Eq (\ref{Control_eq19}), the new state space of the system is formulated in Eq (\ref{Control_eq22}). The system can follow the reference inputs if the designed controller proves the stability of the system. 

\begin{equation}
\label{Control_eq22}
{\begin{array}{c}
\left[ {\begin{array}{c}
\dot{X} \\
   \dot{X}_N 
	\end{array}}\right]= 
	\left[ {\begin{array}{cc}
	A & 0 \\
	-C & 0
	\end{array}}\right]
	\left[ {\begin{array}{c}
		{X} \\
		{X}_N 
	\end{array}}\right]+\left[ {\begin{array}{c}
		B \\
		\Phi
	\end{array}}\right]U + \left[ {\begin{array}{c}
\Phi \\
   I 
	\end{array}}\right]R \\+\left[ {\begin{array}{c} I \\ \Phi \end{array}}\right]D_d  \\
	
	Y = \left[{\begin{array}{cc}
	C & 0 \end{array}}\right] \left[{\begin{array}{c}
	X \\
	X_N
	\end{array}}\right]	
	\end{array}}
\end{equation}
where $\Phi$ is a zero matrix. 

Regarding the acceleration disturbance in the system, the general form of the proposed controller in the time series is given in Eq (\ref{Control_eq23}). 

\begin{equation}
\label{Control_eq23}
u(t) =  k_p e(t) + k_i \int{e(t) dt}+ k_d \dot{e}(t) + k_a \ddot{e}(t)
\end{equation}
where $kp$, $k_i$, $kd$ and $k_a$ are the gain of proposed controller. Then, the MIMO controller is generated by  
\begin{equation}
\label{Control_eq24}
U(s) = \left[ k_p + \frac{k_i}{s} + k_d s + k_a s^2 \right] E(s)
\end{equation}

As seen in Eq (\ref{Control_eq24}), the derivative term is inefficient in the high-frequency domain and can affect the performance of the whole system in a noisy environment. The addition of a derivative filter is proposed to address this issue. Thus, the proposed control is modeled as follows: 

\begin{equation}
\label{Control_eq25}
U(s) = \left[ k_p + \frac{k_i}{s} + k_d \times s L(s) + k_a \times s L(s) \times s L(s) \right] E(s)
\end{equation}
where $L(s)$ is the optimal derivative filter which is formulated as follows:

\begin{equation}
\label{Control_eq20}
L(s) = \frac{N/T}{(N/T) \frac{1}{s}+ 1}
\end{equation}
where $N$ and $T$ are the order of the filter and time constant, respectively. Based on Eq (\ref{Control_eq20}), the transfer function of the optimal derivative filter can be simplified as follows: 

\begin{equation}
\label{Control_eq21}
L(s) = \frac{1}{1 + T_f s}
\end{equation}
where $T_f = T/N$ is the time constant of the optimal derivative filter. Hence, the controller and filter’s parameters can be found by SDSA to minimize the objective function given by Eq (\ref{Control_eq222}).

\begin{equation}
\label{Control_eq222}
f_{obj} = (M_{os}-M_s)^2 - (t_s - t_s)^2
\end{equation}
where $M_{os}$ is the desired maximum overshoot, which is set to $5$ percent; $t_s$, the desired settling time for the system, is $2~sec$. $M_s$ and $t_s$ are the overshoot and settling time for each set of the designed controller. The stability analysis of the system (Eq (\ref{Control_eq15})) is introduced before the simulation results are presented.


\section{Stability Analysis of the Proposed PIDA} \label{Control_sec4}

In this section, the stability of the system, considering the proposed controller is investigated. The following definitions are needed. 

\begin{mydef}
\label{def1}
"Asymptotically stable" is a system around its equilibrium point if it meets the following conditions:

\begin{enumerate}
	\item Given any $\epsilon > 0$, $\exists \delta_{1} > 0$ such that if $\|x(t_0)\| < \delta_1$, then $\|x(t)\| < \epsilon$, $\forall t > t_0$ 
	
	\item $\exists \delta_{2} > 0$ such that if $\|x(t_0)\| < \delta_{2}$, then $x(t) \to 0$ as $t \to \infty$
	
\end{enumerate}

\end{mydef}

\begin{theorem}
$[V(x) = x^T P x, \quad x \in \mathbb{R}^n]$ is a positive definite function if and only if all the eigenvalues of $P$ are positive. 
\end{theorem}

\textit{Proof.}  Since $P$ is symmetric, it can be diagonalized by an orthogonal matrix so $P=U^T D U$  with $U^T U = I$ and $D$ diagonal. Then, if $y = Ux$,
	
\begin{align}
\begin{split}
    V(x) &= x^T P X \\
		 &= x^T U^T D U x \\ 
         &= y^T D y \\
         &= \sum {\lambda}_i |{y_i}|^2 
\end{split}
\end{align}
Thus,
\begin{equation}
    V(x) > 0 \quad \forall x \neq 0  \iff \lambda_i > 0, \quad \forall i 
\end{equation}

\begin{mydef}
\label{def2}
A matrix $P$ is a positive definite if it satisfies $x^T P x > 0 \quad \forall x \neq 0$.
\end{mydef}

Therefore, any positive definite matrix follows the inequality in Eq (\ref{Control_eq27}).

\begin{equation}
\label{Control_eq27}
\lambda_{min} P \|x\|^2 \leq V(x) \leq \lambda_{max} P \|x\|^2
\end{equation}
 
\begin{mydef}
\label{def3}
($V$) is a positive definite function as a candidate Lyapunov function if ($\dot{V}$) has derivative, and it is negative semi-definite function. 
\end{mydef}

\begin{theorem}
\label{theorem1}
If the candidate Lyapunov function (i.e., $V(x) = x^T P x, \quad P>0$) exists for the dynamic system, there is a stable equilibrium point.
\end{theorem}

According to Theorem \ref{theorem1} and the dynamic system defined in Eq (\ref{Control_eq15}), the system in the form of Lyapunov function is as follows: 

\begin{align}
\begin{split}
    \dot{V}(x) & = \dot{x}^TPx + x^TP\dot{x}\\
		& = x^T A^T P x + x^T P A x \\
		& = x^T(A^T P + P A)x \\
		& = -x^T Q x 
\end{split}
\end{align}
where the new notation (see Eq (\ref{Control_eq29})) is introduced to simplify the calculation, it is noted that $Q$ is a symmetric matrix. According to Definition \ref{def3}, $V$ is a Lyapunov function if $Q$ is positive definite (i.e., $Q>0$). Thus, there is a stable equilibrium point which shows the stability of the system around the equilibrium (see Theorem \ref{theorem1}). 

\begin{equation}
\label{Control_eq29}
		A^T P + P A = -Q
\end{equation}

The relationship between $Q$ and $P$ shows that the solution of Eq (\ref{Control_eq29}), called a Lyapunov equation, proves the stability of the system for picking $Q > 0$ if $P$ is a positive definite solution. Thus, there is a unique positive definite solution if all the eigenvalues of $A$ are in the left half-plane. A noisy environment causes the movement of eigenvalues to the right half-plane. Therefore, the system dynamics can intensify instability. This issue raises the cross-coupling among different modes such as roll, pitch, and yaw rate, which are caused by the four rotors. Thus, the derivative term of the proposed controller plays an essential role in maintaining stability. The numerical results show that all eigenvalues of the quadcopter with considering the proposed controller with uncertainties in the environment are in the left half-plane, which it proves that the dynamic system is stable with uncertainties. 

\section{Optimization and Heuristic filter}\label{OptiHeuFil}
In this section, Stochastic Dual Simplex Algorithm (SDSA) and Genetic Filter (GF) are described. First, the optimizaiton algorithm (i.e., SDSA) general setting out is presented. Then, GF, the state estimation module in the proposed controller, is introduced. 

\subsection{Stochastic Dual Simplex Algorithm} \label{sec5}
The heuristic optimization algorithm, named Stochastic Dual Simplex Algorithm (SDSA), is carried out to find the best tuned parameters of the proposed controller. SDSA is the new version of Nelder-Mead simplex algorithm \cite{rao2009engineering}, executing three different operators such as reflection, expansion, and contraction. These operators make dual simplex reshape and move toward the maximum-likelihood regions of the promising area. Each simplex follows the normal rules of simplex, from which the transformed vertices of the general simplex approach are formulated as in Eq (\ref{eq30})-(\ref{eq32}). 
\begin{equation}
\label{eq30}
\textbf{x}_r = (1+\alpha)\bar{\textbf{x}}_0 - \alpha \textbf{x}_h , \quad \alpha > 0
\end{equation}

\begin{equation}
\label{eq31}
\textbf{x}_e = \gamma \textbf{x}_r + (1-\gamma)\bar{\textbf{x}}_0 ,   \quad \gamma > 1
\end{equation}

\begin{equation}
\label{eq32}
\textbf{x}_c = \beta \textbf{x}_h + (1-\beta)\bar{\textbf{x}}_0 , \quad  0 \leq \beta \leq 1
\end{equation}
where $\alpha$, $\gamma$ and $\beta$ are reflection, expansion and contraction coefficients, respectively. During these transformations, the centroid of all vertices excluding the worst point ($\textbf{x}_h$) is $\bar{\textbf{x}}_0$. 

In addition to the movement of dual simplex, a new definition of reflection points is applied to improve the diversity and decrease the probability of local minimum. Therefore, during the \textit{i}-th iteration, the worst vertices of simplexes in search space are replaced by normal distribution directions which are modeled in Eq (\ref{eq33}). 

\begin{equation}\label{eq33}
\overset{*}{\textbf{x}}_{h_s}^{(i)} = \textbf{x}_{h_s}^{(i)} + g^{(i)} {\bar{\textbf{x}}_0}^{(i)}
\end{equation}
where $\overset{*}{\textbf{x}}_{h_s}^{(i)}$ is the new reflected point computed by the worst point of each simplex (${\textbf{x}}_{h_s}^{(i)}$), and $g^{(i)}$ is the normal distribution of the sampled solution in \textit{i}-th iteration and $s$-th simplex. The centroid of all simplexes and the probability density function of the normal distributed simplexes are then expressed in Eq (\ref{eq34}) and  Eq (\ref{eq35}).

\begin{equation}\label{eq34}
\bar{\textbf{x}}_0^{(i)} = \sum_{s=1}^{n_s} {\bar{\textbf{x}}_{0_s}^{(i)}}
\end{equation}

\begin{equation}\label{eq35}
g(\textbf{x}_h|\Sigma) = \frac{1}{\sqrt{2\pi|\Sigma|}}.exp({-\frac{(\textbf{x}_h-\bar{\textbf{x}}_0)^T{\Sigma}^{-1}(\textbf{x}_h-\bar{\textbf{x}}_0)}{2}})
\end{equation}
where $n_s$ and $\Sigma$ are the number of simplexes and covariance matrix of simplexes, respectively.

\par
Reflection makes an action regarding to reflect the worst point, called high, over the centroid $\bar{\textbf{x}}_0$. In this approach, simplex operators utilize the expansion operation to expand the simplex in the reflection direction if the reflected point is better than other points. Nevertheless the reflection output is at least better than the worst, the algorithm carries out the reflection operation with the new worst point again \cite{ZandaviSDSA2019,rao2009engineering}. The contraction is another operation which contracts the simplex while the worst point has the same value as the reflected point. The SDSA pseudocode is presented in Algorithm \ref{SDSA}, and the tuned parameters of SDSA, chosen based on \cite{ZandaviSDSA2019}, are listed in Table \ref{table1}.

\begin{algorithm}
\caption{Stochastic Dual Simplex Algorithm (SDSA)}\label{SDSA}
\begin{algorithmic}

\State \textbf{Initialization}

\State $\quad{\textit{set}} \gets$ \big[$\textit{a}_{max}$,$\alpha_{max}$, $\gamma_{max}$, $\beta_{max}$, $i_{max}$ \big]

\State $\quad{\textbf{x}_0} \gets \textit{random}$

\State \quad{Generate initial $simplexes$}

\State \textbf{Repeat}

\State \quad{Compute Objective Function $(F)$}
\State $\quad{\textbf{x}_h} \gets \textbf{x}_{worst}$

\State \quad\quad{\textbf{while}\;({$\exists \; x_i$})}:
\State \quad\quad\quad\quad{$reflection$}
\State \quad\quad\quad\quad{$expansion$}
\State \quad\quad\quad\quad{$contraction$}
\State \quad\quad{\textbf{end}}

\State $\quad{\textbf{x}_h} \gets \overset{*}{\textbf{x}_h}$

\State \quad{Update $simplexes$}

\State \textbf{Until} Stop condition satisfied.

\end{algorithmic}
\end{algorithm}

\begin{table}
\centering
\caption{Tuned parameters of SDSA}
\label{table1}
\begin{tabular}{l c}
\hline
\textbf{Parameters} & \textbf{Value}  \\ \hline

${a}_{max}$& $10.5907$ \\
${\alpha}_{max}$ & $9.7323$ \\
${\gamma}_{max}$ & $9.9185$ \\
${\beta}_{max}$& $0.4679$ \\
${\textit{i}}_{max}$& $979$ \\

\hline

\end{tabular}
\end{table}

\subsection{Genetic Filter}

In the Genetic Filter (GF), the problem is to estimate the states of a discrete nonlinear dynamic system in a continuous search space. The model is as follows:

\begin{equation}
\textbf{x}_{k} = \textbf{f}_{k}(\textbf{x}_{k-1},\textbf{w}_{k-1})
\end{equation}
Where $k$ is time step, $\textbf{f}_{k}$ is the system model, $\textbf{x}_{k-1}$ is the state vector and $\textbf{w}_{k-1}$ is the process noise corresponding to system uncertainties. Also, the measurement model is considered as

\begin{equation}
\textbf{z}_{k} = \textbf{h}_{k}(\textbf{x}_{k},\textbf{v}_{k})
\end{equation}
where $\textbf{h}_{k}$ is the measurement model and $\textbf{v}_{k}$ is the measurements noise.

Therefore, GF is introduced as a tool for nonlinear systems state estimation based on Genetic Algorithm (GA). GF has two loops. The outer loop generates an initial population that belongs to the first generation every time a new measurement is entered. The inner loop iterates to find the best estimation of the current states, corresponding to the entered measurement. To do this, the inner loop, first propagates the individuals. Then, for each individual, the corresponding output is calculated based on the measurement model. The calculated outputs are compared with the real measurement, and each individual is assigned a cost. The inner loop uses genetic operation such as selection, mutation, and crossover to select new parents and survive the fittest individual and generate new population toward the maximum likelihood regions of the state space and is terminated when the maximum number of iterations (generations) ($\textit{i}_{max}$) is reached. Finally, the state estimation is made using the average of individuals of the last generation (see Algorithm \ref{GF}) . The average of the individuals is calculated and passed as the state estimation. Algorithm \ref{GF} represents pseudo-code of GF.

\begin{algorithm}
\caption[Genetic Filter]{Genetic Filter}\label{GF}
\begin{algorithmic}

\State \textbf{Initialization}

\State $\quad{Set} \gets$ \big[ Max Generation ($\textit{i}_{max}$), Population Size, Mutation Rate \big]

\State \textbf{Repeat}

\State $\quad{\textbf{x}_0} \gets randomize $

\State \quad{Generate initial population}

\State \quad{$i =1$}

\State \quad{\textbf{Repeat}}

\State \quad\quad{Propagate individuals}

\State \quad\quad{Measurement update for each individual}

\State \quad\quad{Compute cost function for each individual}

\State \quad\quad{Update population}

\State \quad\quad\quad{Selection}

\State \quad\quad\quad{Crossover}

\State \quad\quad\quad{Mutation}

\State \quad{$i=i+1$}

\State $\textbf{Until} \gets i \geq i_{max}$

\State $\textbf{Until} \gets \text{measurement is stopped}$

\end{algorithmic}
\end{algorithm}


\section{Numerical Results} \label{Control_sec5}

The numerical simulation is implemented to evaluate the performance of the proposed controller. The model quadcopter was simulated in MATLAB R2016b in a Simulink environment in Windows 10 with an Intel(R) Core(TM)i7-6700 CPU @ 3.4 Hz. The quadcopter parameters are listed in Table \ref{table_Control1}.

\begin{table}[h]
\centering
\caption{Quadcopter Model Parameters}
\label{table_Control1}
\begin{tabular}{c l c c}
\hline
 
\textbf{Parameter} & \textbf{Description} & \textbf{value} & \textbf{Unit} \\ 

\hline

$m$ & Mass & $0.8$ & $kg$ \\
$l$ & Arm length & $0.2$ & $m$ \\
$g$ & Gravity acceleration & $9.81$ & $m/s^2$ \\
$c$ & Force to torque coefficient & $3.00e-5$ & $kg~m^2$ \\
$I_{xx}$ & Body moment of inertia along x-axis & $2.28e-2$ &  $kg m^2$\\
$I_{yy}$ & Body moment of inertia along y-axis& $3.10e-2$ & $kg~m^2$\\
$I_{zz}$ & Body moment of inertia along z-axis& $4.40e-2$ & $kg~m^2$\\
$I_{m}$  & Motor moment of inertia & $8.30e-5$ & $kg~m^2$\\

\hline
\end{tabular}
\end{table}

To begin the simulation and tune the parameters of the proposed controller, the initial state of the quadcopter is at an altitude of $50~m$; and attitude and velocity in different directions are equal to zero. A disturbance, which is modeled as white noise (mean value ($\mu$) is zero and standard deviation ($\sigma$) is one), at time $1~sec$ in the roll channel, is applied to the quadcopter. This disturbance destabilizes the system and locates the eigenvalues of $A$ in the right half-plane. Additionally, the quadrotor is highly sensitive to the noisy environment because of instability and cross-coupling. In this regard, PIDA with a derivative filter, which obviates the noise from the measurement inputs, is designed to respond to this issue and keep the flight stable. 

According to the proposed PIDA with a derivative filter, an additional issue is the tracking of desire inputs, which can be defined as a command to the quadcopter, are another issue that can be addressed by a MIMO controller (i.e., four inputs and four outputs). The proposed controller can be set by four gains and the time constant for each mode/channel.

\begin{figure}
   \centering
    \includegraphics[trim ={4.2cm 8cm 4.5cm 9cm},clip,scale = 0.7]{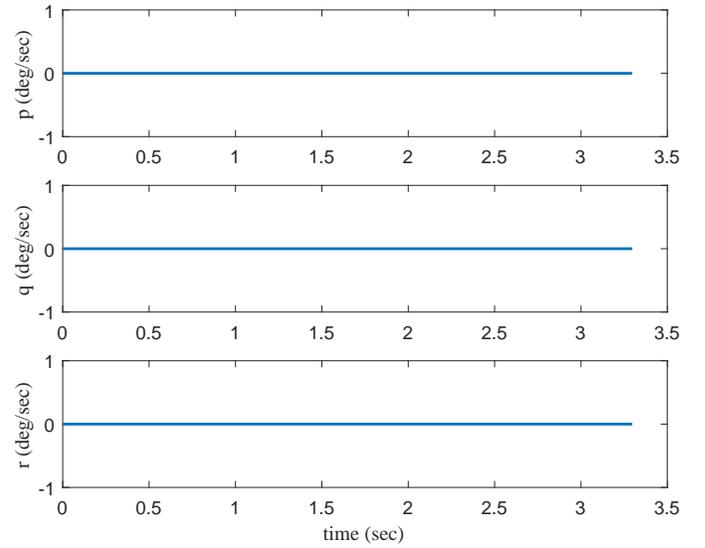}
    \caption{Angular velocity at initial state}
    \label{fig2_PQR_init}
\end{figure}

\begin{figure}
   \centering
    \includegraphics[trim ={4.2cm 8cm 4.5cm 9cm},clip,scale = 0.7]{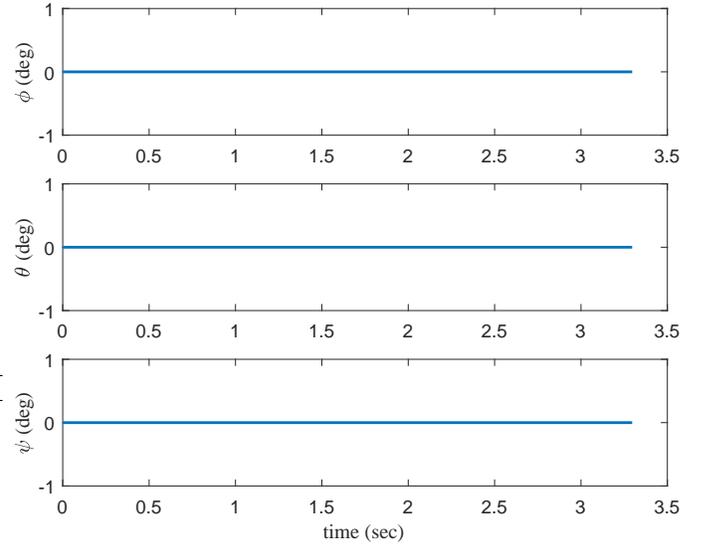}
    \caption{Roll, Pitch and Yaw angle for initial state}
    \label{fig2_PTS_init}
\end{figure}

Figure \ref{fig2_PQR_init} and Fig. \ref{fig2_PTS_init} show the attitude of the modeled quadcopter in the initial state without the noisy environment. It is obvious that the initial state is stable at zero, which is expected to be. 

\begin{figure}
   \centering
    \includegraphics[trim ={4.cm 8cm 4.5cm 9cm},clip,scale = 0.7]{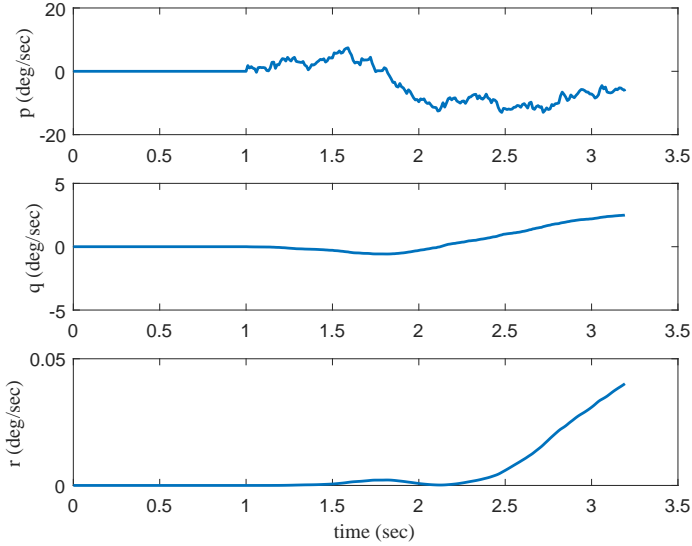}
    \caption{Angular velocity in noisy environment without controller}
    \label{fig2_PQR_dis}
\end{figure}

\begin{figure}
   \centering
    \includegraphics[trim ={4.2cm 8cm 4.5cm 9cm},clip,scale = 0.7]{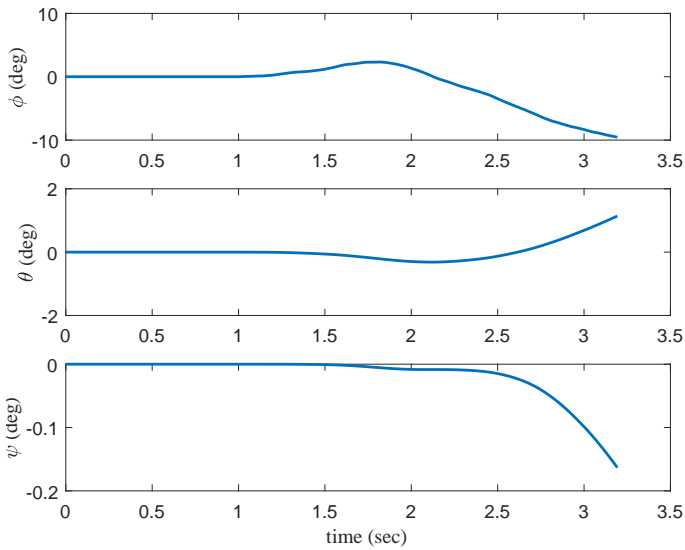}
    \caption{Roll, Pitch and Yaw angle for noisy environment without controller}
    \label{fig2_PTS_dis}
\end{figure}

A disturbance, which is modeled as white noise (mean value ($\mu$) is zero, and standard deviation ($\sigma$) is one), at time $1~sec$ in roll channel is applied to the quadcopter. As seen in Figs. \ref{fig2_PQR_dis} and \ref{fig2_PTS_dis}, this disturbance renders the system unstable. Hence, the quadrotor is highly sensitive in the noisy environment. In this regard, PIDA with a derivative filter, which obviates the noise from the measurement inputs, is designed to respond to this issue and keep the flight stable. 

To tune the parameters of PIDA, the complex commands that enable coupling among different modes of the modeled quadcopter are used to evaluate the performance of the designed controller. New command angles are provided by a step function with $2~sec$ delay time in the simulation environment, where $\phi = -5^{\circ}$, $\theta = 10 ^{\circ}$, $\psi = 30 ^{\circ}$ and with altitude starting from $50~m$ and stabling at $20~m$. Note that noisy measurements have been considered for this simulation, and are modeled as white noise. 

\begin{figure}
    \centering
    \includegraphics[trim ={4.2cm 8.6cm 4.5cm 8.5cm},clip,scale = 0.7]{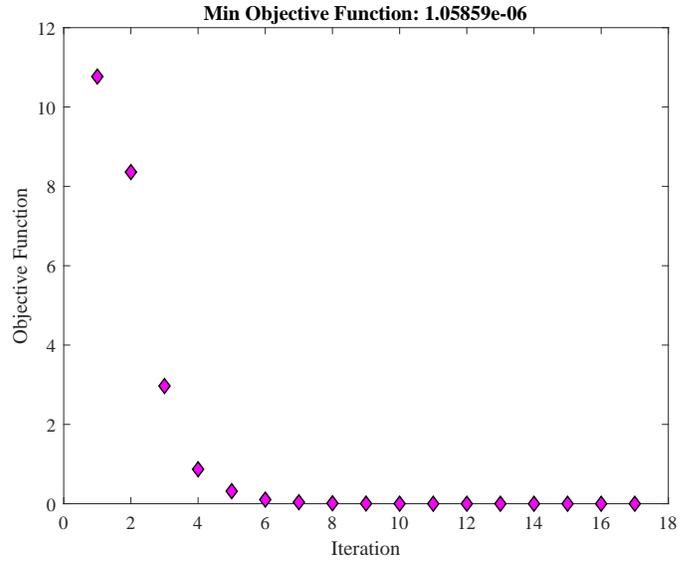}
    \caption{Performance of SDSA versus iteration }
    \label{fig1&1}
\end{figure}

The parameters of controllers are tuned using SDSA \cite{ZandaviSDSA2019}, and the convergence graph is shown in Fig. \ref{fig1&1}. The SDSA is applied to the objective function introduced in Eq (\ref{Control_eq222}). Table \ref{table_Control2} represents the best fit set of parameters for different modes/channels within the noisy environment.  

\begin{table}[h]
\centering
\caption{Controller Parameters for Altitude and Attitude}
\label{table_Control2}
\begin{tabular}{ccccc}
\hline

\multirow{2}{*}{\textbf{Controller Parameter}} & \multicolumn{4}{c}{\textbf{Channel}}\\

\cline{2-5}

& Roll & Pitch & Yaw & Altitude \\

\hline

$k_i$ & $0.1436$ & $3.6869$ & $0.0437$ & $1.00$\\
$k_d$ & $6.5097$ & $21.2743$& $29.9872$& $11.4676$ \\
$k_a$ & $0.5772$ & $0.3429$ & $23.5238$& $7.5114$ \\
$T_f$ & $0.0437$ & $0.0331$ & $0.0117$ & $0.3752$ \\

\hline
\end{tabular}
\end{table}

\begin{figure}
    \centering
    \includegraphics[trim ={4.2cm 8cm 4.5cm 9cm},clip,scale = 0.7]{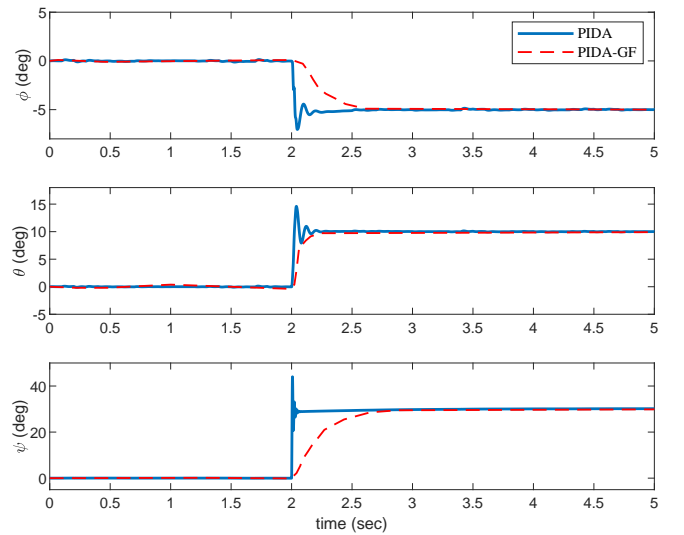}
    \caption{Controller response to the change of Roll, Pitch and Yaw angle for PIDA and PIDA associated with GF}
    \label{fig_Control2}
\end{figure}

\begin{figure}
    \centering
			\includegraphics[trim ={4.0cm 8cm 4.5cm 8cm},clip,scale = 0.7]{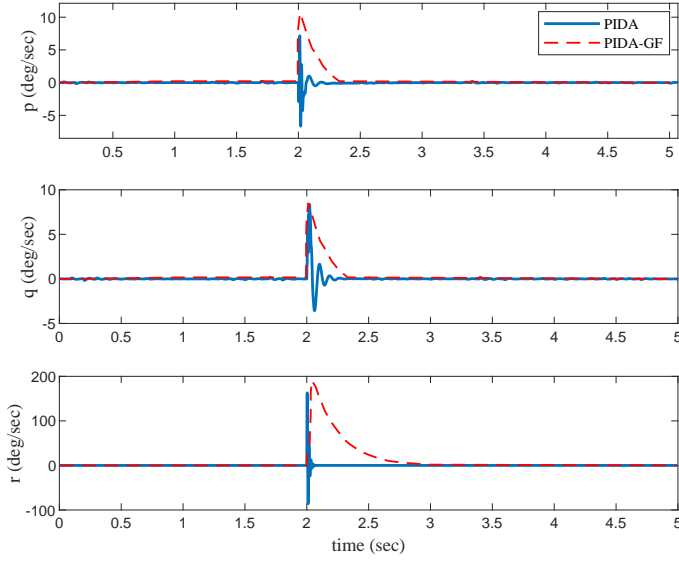}
    \caption{Controller response to the change of Angular velocity PIDA and PIDA associated with GF}
    \label{fig_Control3}
\end{figure}

\begin{figure}
    \centering
			\includegraphics[trim ={4.2cm 8cm 4.5cm 9cm},clip,scale = 0.7]{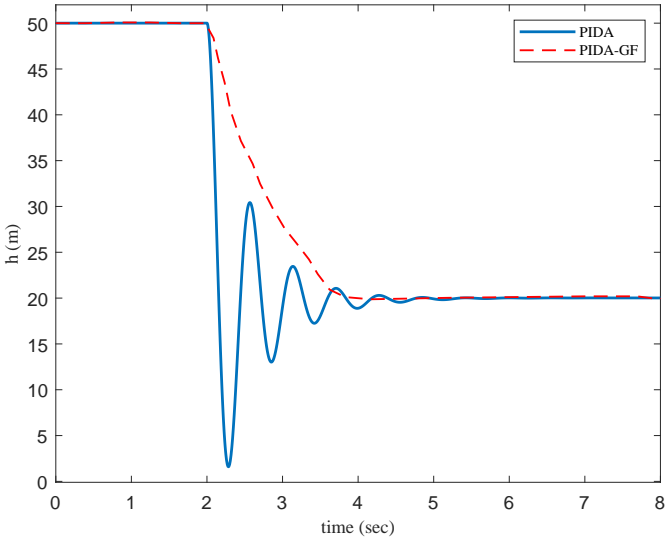}
    \caption{Controller response to the change of Altitude PIDA and PIDA associated with GF}
    \label{fig_Control4}
\end{figure}

The delay time causes missing measurements in the noisy environment, so the robust heuristic filters is required. GF, as robust filter for the dynamic system, plays an important role to keep the flight stable by accurate estimations and removing noise. Figures \ref{fig_Control2}–\ref{fig_Control4} show that the presented controller with GF can adequately respond to and track the reference commands in the noisy environment.

Having PIDA tuned under environmental uncertainties, a particular spiral trajectory is introduced to evaluate the performance of the proposed control in the noisy environment. The evaluated trajectory is modeled in Eq (\ref{Eq:sprialTraj}). 

\begin{equation}
\label{Eq:sprialTraj}
    {\begin{array}{c}
    x = 2sin(3 \omega t) + 2 cos(\omega t)\\
    y = 2 sin(\omega t) + 2 cos(3 \omega t)\\
    z = 0.3 t
	\end{array}}
\end{equation}
where $x$, $y$, and $z$ are the reference trajectory. $\omega$ is the period spiral trajectory, which is set to $\omega = 1/2\pi$. $t$ is the flight time between $0~sec$ and $60~sec$. 

\begin{figure}
    \centering
    \includegraphics[scale = 0.55]{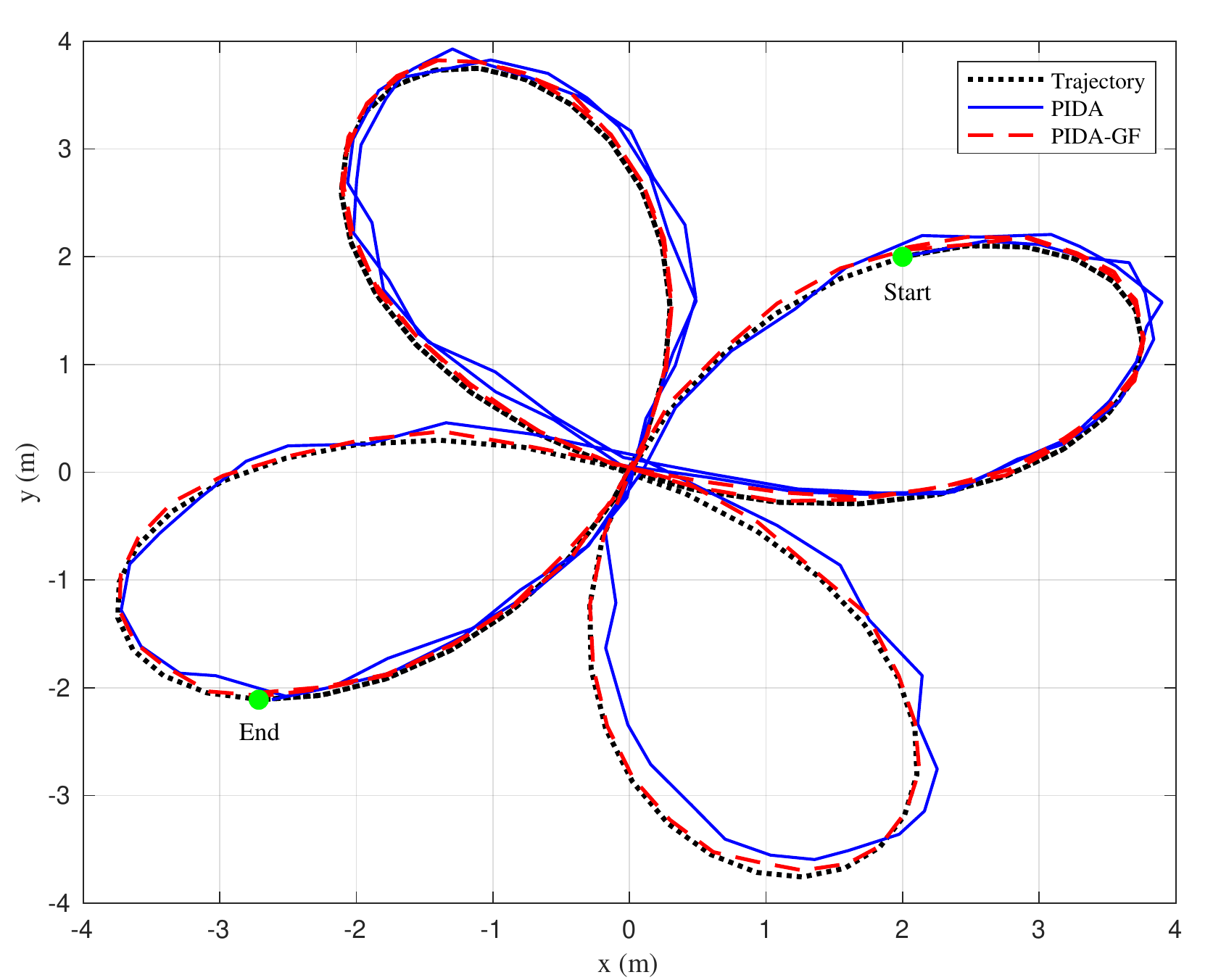}
    \caption{2D trajectory of drone under uncertainties for PIDA and PIDA-GF}
    \label{fig:2dTrajectory}
\end{figure}

\begin{figure}
    \centering
    \includegraphics[scale = 0.55]{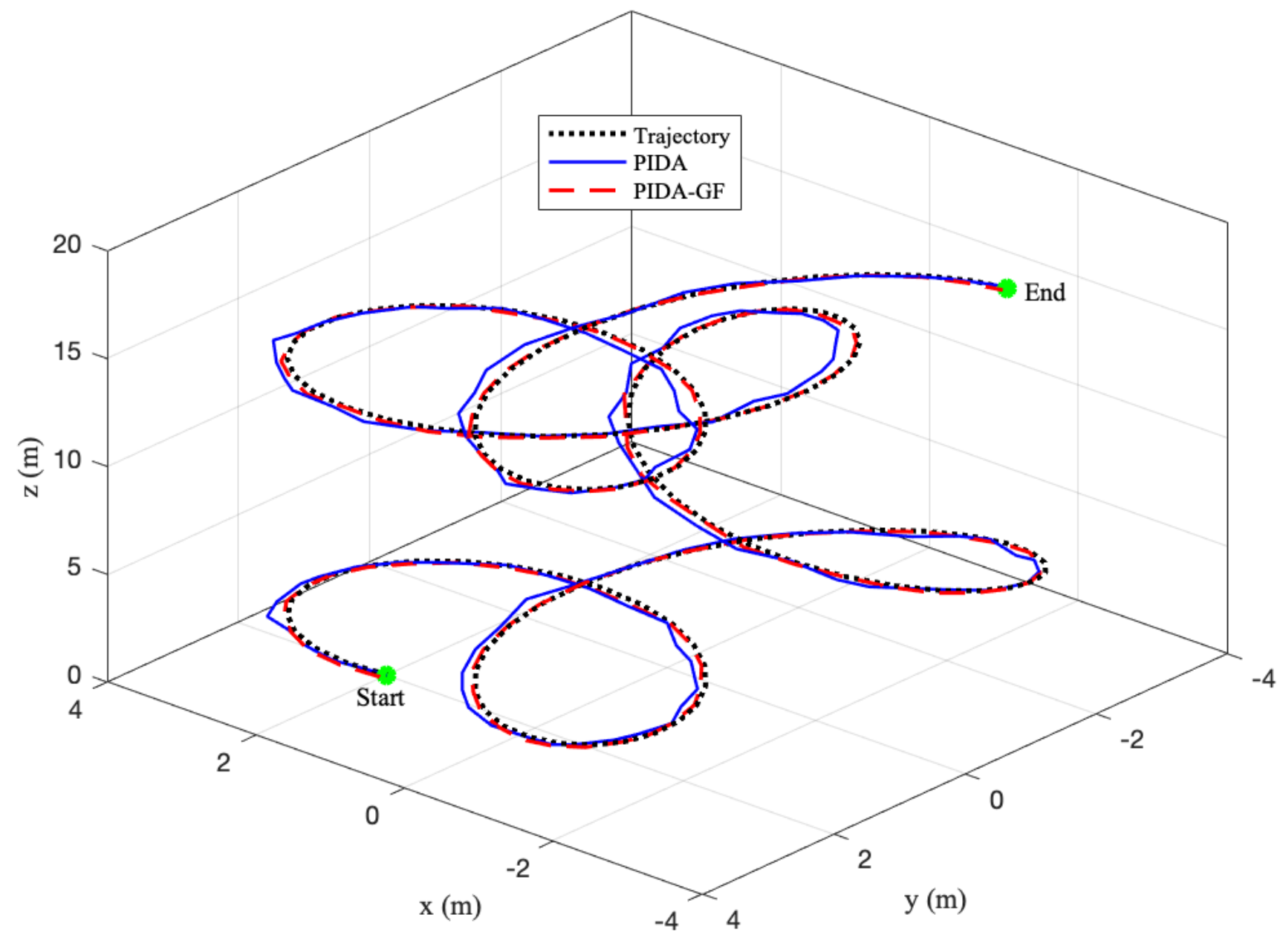}
    \caption{3D trajectory of drone under uncertainties for PIDA and PIDA-GF}
    \label{fig:3dTrajectory}
\end{figure}

Figures \ref{fig:2dTrajectory} and \ref{fig:3dTrajectory} demonstrate the drone trajectory for both PIDA and PIDA-GF. These figures show that PIDA associated with GF can boost the performance of the proposed controller in the complex dynamics. Thus, the proposed controller is able to have a great response in the spiral trajectory.  

\begin{figure}
    \centering
    \includegraphics[scale = 0.50]{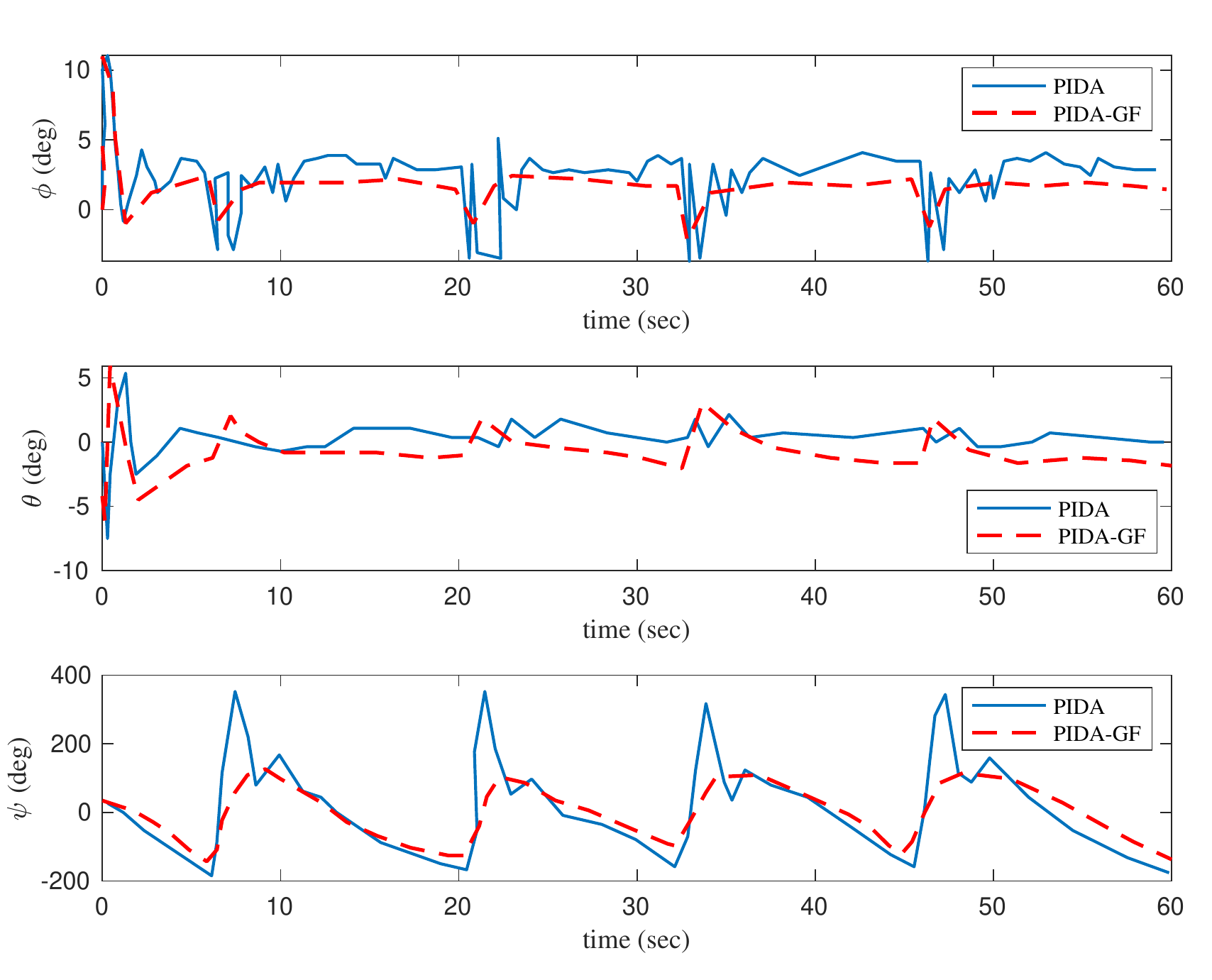}
    \caption{Drone responses in the spiral trajectory}
    \label{fig:PTSSpiral}
\end{figure}

Accordingly, the drone movement in cooperation with PIDA-GF can guarantee a smooth and stable flight while the drone is flying in the noisy environment. Figure \ref{fig:PTSSpiral} shows the drone responses in the reference path. As seen in Fig. \ref{fig:PTSSpiral}, PIDA responses fluctuated around the points when the drone turns in the spiral trajectory. This is due to the noise in the environment. Accordingly, GF is able to reduce the effect of noise in the environment; thereby, the smooth flight is performed by PIDA-GF. 

Consequently, these figures and the reference trajectory demonstrate that the proposed controller associated with GF is powerful enough to control the spiral path of drone.

As the simulation results demonstrate, not only is the quadcopter capable of stable flight, but the proposed controller associated with GF also provides a smooth flight. This smoothness and handling are acquired because of having the robust filter, GF, in the dynamic movement. As shown, PIDA provides a fast and stable flight, but applying GF for integrated estimation of the observed states and parameters in both attitude and altitude boosts performance in the quadcopter. 


\section{Conclusion} \label{Control_sec6}

This paper has proposed a new Proportional-Integral-Derivative-Accelerated (PIDA) controller with a derivative filter to improve flight stability for a quadcopter and considers the noisy environment. The mathematical model considering non-linearity, uncertainties, and coupling was derived from an accurate model with a high level of fidelity. In the indoor environment and as the critical features in the proposed controller, overshoot and settling time were limited during the operation of the dynamic system (drone). The noisy environment causes the movement of poles to the right half-plane. Thus, system dynamics intensify instability. This issue raises the cross-coupling among different modes, such as roll, pitch, and yaw, which were generated by the four rotors. The proposed controller and heuristic Genetic Filter (GF) addressed these challenges. Moreover, the derivative term of the proposed controller assists the dynamic system to recover its stability. The controller gains were optimized before performing the mission. In this regard, the tuning of the proposed controller was performed by Stochastic Dual Simplex Algorithm (SDSA). The simulation results show that the proposed PIDA controller associated with GF was capable of supporting outstanding performance in tracking the desired point despite disturbances.


\bibliographystyle{IEEEtran}
\bibliography{Refs}

\end{document}